# From Artefact to Insight: Efficient Low-Rank Adaptation of BrushNet for Scanning Probe Microscopy Image Restoration

*Ziwei Wei, Yao Shen, Wanheng Lu, Ghim Wei Ho, and Kaiyang Zeng\**

Z. Wei, Prof. K. Zeng

Department of Mechanical Engineering, National University of Singapore,

9 Engineering Drive 1, 117576, Singapore

Prof. K. Zeng

NUS Research Institute (NUSRI), No. 377 Linquan Street,

Suzhou Industrial Park, 215123, China

Y. Shen

School of Computer Science and Technology, Tongji University,

4800 Caoan Road, Shanghai, 201804, China

Dr. W. Lu, Prof. G. W. Ho

Department of Electrical and Computer Engineering, National University of Singapore,

4 Engineering Drive 3, Singapore, 117583

*Corresponding author. E-mail: mpezk@nus.edu.sg

KEYWORDS

scanning-probe microscopy; image in-painting; diffusion models; low-rank adaptation; artefact removal; generative restoration

ABSTRACT

Scanning Probe Microscopy (SPM) offers nanoscale resolution but is frequently marred by structured artefacts such as line-scan dropout, gain-induced noise, tip convolution, and phase hops. While most available methods treat SPM artefact removal as isolated denoising or interpolation tasks, the generative inpainting perspective remains largely unexplored. In this work, we introduce a diffusion-based inpainting framework tailored to scientific grayscale imagery. By fine-tuning < 0.2 % of BrushNet weights with rank-constrained low-rank adaptation (LoRA), we adapt a pretrained diffusion model using only 7390 artefact – clean pairs distilled from 739 experimental scans. On our forthcoming public SPM-InpBench benchmark, the LoRA-enhanced model lifts the Peak Signal-to-Noise Ratio (or PSNR) by 6.61 dB and halves the Learned Perceptual Image Patch Similarity or LPIPS relative to zero-shot inference, while matching or slightly surpassing the accuracy of full retraining, trainable on a single GPU instead of four high-memory cards. The approach generalizes across various SPM image channels including height, amplitude and phase, faithfully restores subtle structural details, and suppresses hallucination artefacts inherited from natural-image priors. This lightweight framework enables efficient, scalable recovery of irreplaceable SPM images and paves the way for a broader diffusion-model adoption in nanoscopic imaging analysis.

**INTRODUCTION**

Scanning Probe Microscopy (SPM) has been indispensable for resolving surface processes in many research fields, such as functional materials, catalysis, energy storage materials, biomolecular science and many others.[1–4] However the insight it offered hinges on a deceptively fragile requirement: the final image usually needs to be free of scanning artefacts, so that the detail structural or functional features can be viewed and analysed accurately. On the other hand, mis-tuned feedback loops generally imprint many noise or artefacts such as high-frequency stripes, abrupt phase hops seed speckled islands, and asymmetric or worn tips may lead to directional tails that elongate real features, and piezo creep may also introduce global warping or slope.[5–7] A concise taxonomy of these defects, their physical causes and visual signatures is summarized in Table 1. For novices, some defects are hard to spot, and even seasoned operators rarely obtain a "defect-free" frame on the first attempt. Fragile or single-use specimens typically cannot tolerate repeated scanning; and in time-series experiments, even a single corrupted frame can force the entire sequence to restart. Such scenarios not only squander valuable sample materials and instrument time but also highlight the pressing need for a robust artefact-removal solution.

Existing counter-measures remain fragmented. Hardware solutions, such as high speed scanners suppress feedback-induced dropout, while active drift-compensation stages attenuate thermal creep, are effective yet costly and sometime not function against random events.[8,9] Emerging reinforcement-learning and machine-learning controllers dynamically retune imaging parameters, but these methods typically require substantial adjustments in the existing systems[10] or focus narrowly on specific imaging modes[11] or artefact types.[12] Software approaches range from classical non-learning inpainting methods[13–17] to compressed sensing[18] and blind-tip deconvolution,[19–21] each excelling under narrow assumptions but deteriorating when multiple classes of defect coexist. What still lacking is an automatic, data-driven tool that "can see through" the noise with the intuition of an expert while preserving nanometer-scale fidelity.

Generative diffusion models are a natural fit for such in-painting tasks, yet three obstacles have so far blocked their adoption in the SPM image analysis: (i) pre-training on Internet photos that diverge sharply from grayscale height maps; (ii) the community's demand for pixel-accurate continuity rather than merely plausible texture; and (iii) the impracticality of collecting millions of annotated SPM images.

This study therefore aims to bridge this gap by adapting BrushNet,[22] a dual-branch diffusion architecture that decouples masked image features from generative noise, to the domain of SPM image analysis. Instead of retraining all weights, in this work, we apply low-rank adaptation (LoRA) method,[23] fine-tuning < 0.2 % of parameters on just a few hundred carefully curated scans. LoRA is selected over heavier schemes such as DreamBooth[24] or ControlNet[25] because it preserves the pretrained prior, avoids explicit structural conditioning, and runs on commodity hardware (comparative details are in the Discussion Section of this paper).

In this study, masks are generated with tailored Segment Anything Method (SAM) and pruned manually; an ignore-region loss discards uncertain pixels so even lightly contaminated images can contribute to learning. When evaluated on SPM InpBench, a benchmark we build and available publicly, LoRA-tuned BrushNet boosts PSNR (Peak-Signal-to-Noise Ratio) by 6.61 dB and halves perceptual error versus zero-shot inference, while cutting the hardware footprint from four high-memory GPUs (17–19 GB each) to a single commodity card (7–21 GB). The model generalizes across height, amplitude and phase channels and can exploit phase-channel anomalies to flag corresponding artefacts in height maps that would otherwise evade visual inspection.

To our knowledge, in the field of SPM image analysis, by casting artefact removal as a generalized in-painting problem and delivering the first diffusion-based solution tailored to scientific grayscale imagery, this approach lowers the learning curve for newcomers and salvages irreplaceable datasets for experts.

The rest of this paper describes the dataset, model architecture and training scheme (Methods Section); performance quantitatively evaluation, and demonstrating the applications on real experimental data (Results Section); as well as the comparision with the traditional approaches, discusseing current limitations and directions for future improvement (Discussion Section).

**RESULTS**

**A. SPM-Artefact Taxonomy.**

Table 1 categorizes prevalent SPM artifacts by their physical origins, visual impacts, and identifies targeted in this work. Specifically, we address structured degradations difficult to resolve via conventional filtering or calibration methods, including line-scan dropout, gain-related noise, tip-induced directional tailing, and phase-hop artifacts. In contrast, simpler issues such as periodic stripe noise or global thermal drift can often be mitigated effectively using established signal-processing or calibration approaches and thus not discussed in this paper.

**Table 1**. Taxonomy of frequent SPM artifacts.

| Artefact class | Physical origin | Visual impact | Addressed? |
|---|---|---|---|
| Line-scan dropout | Tip disengagement | Single bright/dark streak | Yes |
| Gain-related noise | Feedback saturation | High-freq band(s) | Yes |
| Tip-induced tailing | Asymmetric/blunt tip | Elongated duplicate edges | Yes |
| Phase-hop artefact | Abrupt force change | Invalid patches | Yes |
| Periodic stripe noise | External interference | Regular background bands | No |
| Trace–retrace mismatch | Scanner hysteresis | Offset forward/reverse trace | No |
| Thermal drift / tilt | Piezo creep | Global slope/warp | No |
| Quantization noise | Low-res ADC | Stair-step flats | No |

Four dominant artefact families are encountered in the dataset for ths work (Figure 1). Each sub-panel shows the representative micrograph; furthermore, additional variants are illustrated in Supplementary Figure S1 (Support Information). These also form the basis for all subsequent quantitative (Section 2.3) and qualitative (Section 2.4) analysis.

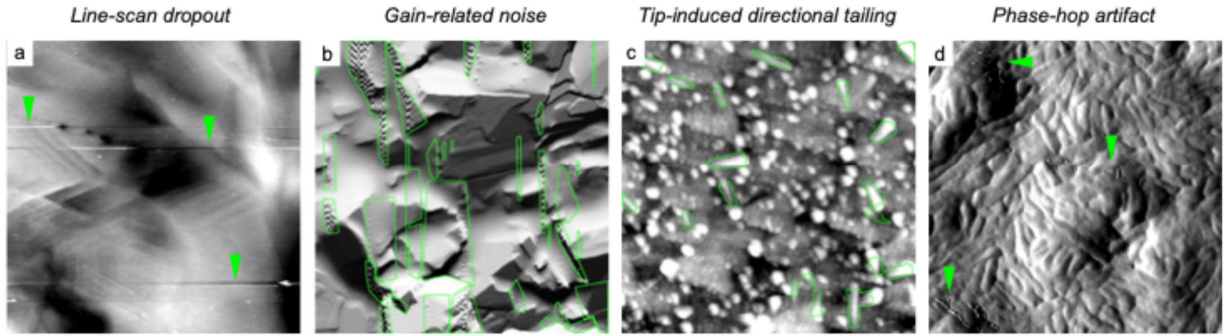

**Figure 1**. Representative examples of the four classes of artefact targeted in this study: (a) Line-scan dropout: a single horizontal streak caused by transient tip disengagement; (b) Gain-related noise: high-frequency intensity fluctuation in low-signal regions due to excessive feedback gain; (c) Tip-induced directional tailing: elongated "tails" produced by asymmetric or blunt tips, or by tip dragging on soft samples; and (d) Phase-hop artefact: patches of invalid data generated when the cantilever during tapping-mode operation momentarily loses stable phase lock.

**B. Metrics Rationale.**

All evaluation metrics are computed within the masked region to assess restoration quality only where the model operates. We adopt following evaluation parameters: Peak Signal-to-Noise Ratio (PSNR[26]) and Structural Similarity Index (SSIM[27]) for quantifying pixel-level fidelity, and Learned Perceptual Image Patch Similarity (LPIPS[28]) for measuring perceptual similarity based on deep feature embeddings. Furthermore, Mean Squared Error (MSE[29]) is also computed for completeness, although its sensitivity to rare outliers and limited perceptual relevance on monochrome nano-scale data make it a less informative metric. Formal definitions of those parameters are provided in Section 5.3 as well as Equations (S1) – (S4) in the Supplementary Information (SI).

## C. Benchmark-based Quantitative Evaluation.

Next, we benchmark inpainting performance across three different settings: i.e., direct test-only inference, full BrushNet retraining, and LoRA-based fine-tuning. Quantitative results are shown in Tables 2 and 3, with representative visual examples in Figure 2. Section 2.4 and Figure S2 in Supplementary Information (SI) present more case studies on real-world experimental artifacts beyond the current *SPM InpBench* datasets. Additional restoration videos can be found in the Supplementary Videos which are organized according to material class: SV1: Crystalline Inorganic Solids (49 images); SV2: Metals & Alloys (25 images); SV3: Layered 2D & Anisotropic Materials (15 images); SV4: Polymers & Polymer Composites (37 images); SV5: Hybrid/Composite and MOF Systems (36 images), and SV6: Biological & Calibration References (67 images).

Retraining the full BrushNet backbone significantly improves accuracy over the test-only inference but involves optimizing all 619 million parameters, this typically requires the computation power with at least four high-memory GPUs. By contrast, the LoRA variant used in this study introduces lightweight rank-8 adapters into the selected layers of BrushNet, training less than 0.2% of total parameters. These adapters guide the frozen pretrained backbone to adapt efficiently, preserving prior knowledge while minimizing overfitting risk. This is especially advantageous because: (i) the SPM corpus is $\sim 600\times$ smaller than the BrushNet's original 3.2 TB dataset, making full retraining prone to overfitting and catastrophic forgetting; and (ii) most laboratories can afford a single overnight run on an 8 GB GPU or similar, but not multi-GPU training over several days. This work shows that LoRA can achieve comparable and, in several cases, superior inpainting quality, while at the same time, cutting computational cost by orders of magnitude.

**C1. Core metrics on SPM-InpBench.** Table 2 summarizes human-aligned perceptual quality (LPIPS), pixel fidelity (PSNR, MSE), and structural similarity (SSIM), on *SPM InpBench*, which contains 415 ground-truth frames withheld from training. Genearlly speaking, lower LPIPS and MSE values indicate better reconstruction quality; higher PSNR and SSIM values reflect greater visual and structural accuracy. LoRA fine-tuning and full BrushNet retraining are compared against direct test-time inference. All scores are computed inside the green mask regions so that we assess only the region the model must hallucinate.

**Table 2.** Quantitative evaluation on *SPM-InpBench*.

| Models | $LPIPS_{\times 10^3}$ ↓ | PSNR↑ | $MSE_{\times 10^3}$ ↓ | $SSIM_{\times 10^2}$ ↑ | Best step |
|---|---|---|---|---|---|
| Test-only | 17.36 | 15.26 | 46.62 | 96.59 | — |
| Full train | 7.56 | 21.84 | 15.96 | 97.81 | 600 |
| LoRA | 7.64 | 21.87 | 15.74 | 97.80 | 5000 |

*Accuracy gains.* Both full BrushNet retraining and LoRA fine-tuning markedly outperform test-only inference across all metrics. Full training improves PSNR by +6.58 dB and reduces LPIPS and MSE by over 55% and 65%, respectively, confirming that adapting the model to the SPM domain is essential. SSIM also increases from 96.59 to 97.81, reflecting enhanced structural consistency.

*LoRA advantage.* Compared to full retraining, LoRA achieves a similar PSNR performance but provides a more stable and generalized approach. The full retraining process achieves peak performance rapidly, reaching an optimal PSNR of 21.84 dB after approximately 600 steps, but it overfits quickly beyond 800 steps (PSNR: 21.84 dB to 20.30 dB). In contrast, LoRA reaches peak performance at around 5000 steps and maintains high performance, with only marginal degradation at 10,000 steps (PSNR: 21.87 dB to 21.61 dB). This difference highlights the advantages of LoRA in preventing overfitting, particularly given the relatively small dataset of 7390 image pairs.

**C2. Efficiency analysis.** Table 3 summarises resource usage. LoRA keeps the trainable footprint at $7.5 \times 10^5$ parameters, a 99.88 % reduction comparison with what used by full retraining. LoRA proves to be a more computationally efficient solution, especially in environments with limited resources or when utilizing single-GPU setups. By increasing the batch size in the throughput-saturated regime, LoRA can achieve results comparable to or even better than what can be achieved by using Full BrushNet over longer training periods, offering a better balance between performance and efficiency. This flexibility allows researchers to select the most suitable training approach based on available computational resources. However, Full BrushNet is highly susceptible to overfitting and catastrophic forgetting, particularly when working with small training datasets. In the scientific domains such as SPM image analysis and recostruction, where data is sparse and material heterogeneity demands strong generalization, LoRA's ability to preserve pretrained priors while fine-tuning specific parameters makes it the more robust choice. Therefore, this work recommends LoRA as the preferred training method for such applications.

**Table 3.** Efficiency comparison of full BrushNet training versus LoRA fine-tuning.

| Models | Trainable params | Time (to best) | #GPU × VRAM | PSNR ↑ |
| --- | --- | --- | --- | --- |
| Full (EB-4[a]) | $6.19 \times 10^8$ | 0.75 h | 4 × 17 GB | 20.94 |
| LoRA (EB-4) | $7.48 \times 10^5$ | 4.50 h | 1 × 7 GB | 20.27 |
| Full (EB-12) | $6.19 \times 10^8$ | 0.15 h | 4 × 19 GB | 21.84 |
| LoRA (EB-12) | $7.48 \times 10^5$ | 3.50 h | 1 × 21 GB | 21.87 |

a) Effective batch (EB) = #GPU × micro-batch × accumulation. EB-4 corresponds to an effective batch of 4, and EB-12 corresponds to an effective batch of 12.

**C3. Visual diagnostics.** Figure 2 illustrates five representative mask archetypes, progressing systematically from routine defects to extreme stress tests. The green overlays mark the regions removed from the input and presented as blanked-out areas to the model for restoration; ground-truth references are shown on the far right of the figure.

Images on Row (a) show small isolated gap, which is the typical defect in SPM images. A single, compact region (< 5% of the frame) removed from a height-ridge. While all methods restore the ridge topology, LoRA achieves noticeably sharper edge transitions than that of the full fine-tuning, corroborating its superior PSNR and SSIM metrics.

Row (b) shows a broad, homogeneous dropout or large area defects in a simple context. A contiguous, featureless area is missing. The test-only model leaves a residual depression; both adaptive approaches restore a flat, continuous surface. However, full fine-tuning exhibits slight brightness bias, suggestive of mild over-adaptation. In contrast, LoRA maintains faithful quantitative alignment with the ground truth.

Row (c) shows a multi-line scan dropout, which is a classical line-scan artefact in the SPM images. Multiple contiguous scan lines are removed, creating a scenario that defies simple local reconstruction. Here, full fine-tuning displays noticeable intensity shifts and blurred periodic features, which is clear indicators of overfitting due to limited data. LoRA, on the other hand, leveraging low-rank parameter constraints, accurately restores and preserves contrast fidelity.

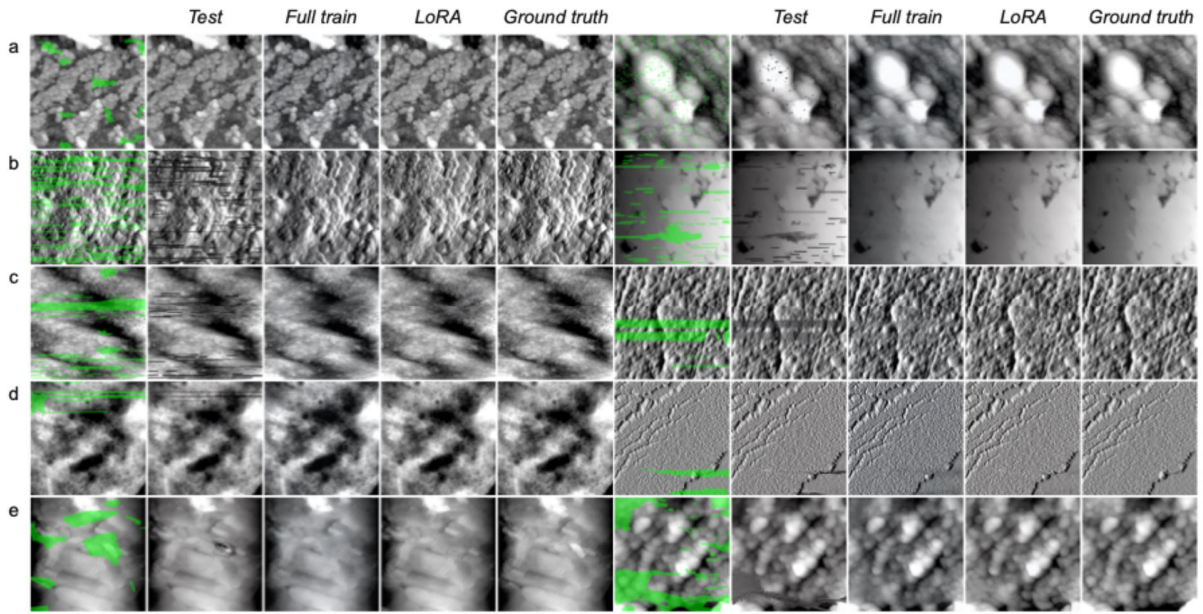

**Figure 2.** Representative inpainting results across five mask archetypes. Each group shows the corrupted input (Test), results from full BrushNet retraining and LoRA fine-tuning, and the ground truth. Green overlays indicate masked regions shown to the model. Rows illustrate progressively harder inpainting scenarios: (a) small isolated gap; (b) broad homogeneous dropout; (c) multi-line scan dropout; (d) directional tailing in textured regions; and (e) near-tabula-rasa masks.

Row (d) images show the dense directional tailing in texture-rich regions, i.e., complex background. Sub-micron anisotropic striations require preservation of precise local statistical relationships. Full fine-tuning smooths fine texture details and slightly alters the grayscale distribution. In contrast, LoRA precisely replicates striation orientations and native contrasts, underscoring its ability to preserve texture patterns learned during extensive pre-training, due to the minimal parameter updates preventing catastrophic forgetting.

Row (e) images are using a near-tabula-rasa mask, i.e., "free-painting" scenario. Here, key local structures (e.g., the left-hand protrusion) are fully occluded, leaving the network to reconstruct content without structural cues. Because such large-area masking is rare in practice, perfect registration with ground truth is neither expected nor feasible; the goal in this study is therefore simply to test whether outputs can remain consistent with the ground truth. The inference-only baseline hallucinates natural-image artefacts, i.e., an eye-like feature on the left and a leaf-like motif on the right, whereas both adapted models respect true topography. LoRA further sharpens feature boundaries, highlighting its robustness under sparse context and its ability to suppress the natural-image bias of the pretrained weights.

Full fine-tuning modifies all 619 million parameters using a limited (∼5 GB) corpus, which is approximately 600-fold smaller than its pre-training set. Consequently, rare artifacts disproportionately influence the training gradients, resulting in noticeable intensity shifts and texture degradation. In contrast, LoRA restricts learning to less than 0.2% of parameters, allowing the frozen backbone to provide a stable global prior while dedicated low-rank adapters selectively capture SPM-specific characteristics. This effective separation significantly reduces overfitting, yielding enhanced restoration fidelity in challenging scenarios (Figs.2 rows (c) – (d)) and mitigating the hallucination of irrelevant natural-image features (Figs.2 row (e)).

To avoid distracting from the core evaluation, we do not elaborate scan sizes or imaging conditions here; these are detailed in Supplementary Section 3 (SI). Additional qualitative comparisons can also be found in Supplementary Figure S2 (SI).

**D. Qualitative Restoration on Real-World SPM Data.**

To confirm that our benchmark results translate into tangible benefits under realistic SPM conditions, we next apply the optimized LoRA-enhanced BrushNet to authentic experimental datasets. Specifically, we illustrate how this model effectively handles prevalent real-world artefacts: (i) line-scan dropout; (ii) tip-induced directional tailing; and (iii) phase-hop events. Those artefacts frequently disrupt actual nanoscale imaging workflows, and many time, repeated scanning are needed in order to get "defects-free" images. These real-data demonstrations validate both the quantitative improvements achieved during benchmarking and highlight the practical applicability of our approach in genuine research scenarios, thereby bridging the gap between theoretical metrics and experimental reliability.

**D1. Line-scan artifacts.** The first category of artifacts is the pervasive line-level corruption often encountered in the SPM images. These artifacts appear as repetitive horizontal lines or bands across the image, typically caused by electronic or feedback instabilities during the data acquisition. For example, improper grounding or an excessively high feedback gain can imprint a persistent striping pattern on the data. Line noise is in fact one of the most notorious issues in the SPM imaging, arising from the line-by-line scanning processes, any abrupt tip perturbation or data dropout on a scan line can result in a streak that cannot be averaged out. Despite their conspicuous appearance, experienced microscopists can often still discern the underlying sample features, as the overall structural trends remain visible beneath the noise. The current model is designed to do the same: remove the overlay while preserving the genuine morphology of the specimen. Figures 3(a–c) exemplify stripe and gain noise on three different materials: a metal–organic framework or MOF (2-amino-methyl manganese hypophosphite, $CH_3–2NH_2MnH_2PO_2–3$); a PVDF (Polyvinylidene Fluoride)-Ag composite; and a PZN–PT ($Pb(Zn_{1/3}Nb_{2/3})O_3$-$PbTiO_3$) ferroelectric crystal, respectively. In each raw images, pronounced line artifacts are superimposed on the true surface texture. Notably, the severity of the noise increases from Figure 3(a) to Figure 3(c); by panel (c), broad regions are affected by scanline-wise signal distortions due to drifting feedback gain, which is a challenging scenario for conventional interpolation-based corrections. The model's output successfully excises the line noise in all cases, revealing a clean rendition of the material's nanostructure. Even in the extreme case showed in the Figure 3(c), the large-area banding is eliminated without introducing discontinuities, and the subtle granular details of the sample are retained. Another variant of this artifact is illustrated in Figures 3(d) – (f), which show an irregular, large-area striping pattern in an LAGP ($Li_{1+x}Al_xGe_{2-x}(PO_4)_3$) solid electrolyte sample. Here the noise does not form neat lines but rather blotchy regions of spurious contrast, likely stemming from erratic gain changes or external interference. Traditional approaches struggle with such non-uniform artifacts, whereas the model demonstrates a strong inpainting capability, it fills in the corrupted regions in panels (d)–(f) while faithfully preserving all real morphological features (e.g., the grain boundaries in the LAGP). This indicates that this model can distinguish and remove even complex stripe/gain artifacts without damaging valid image information.

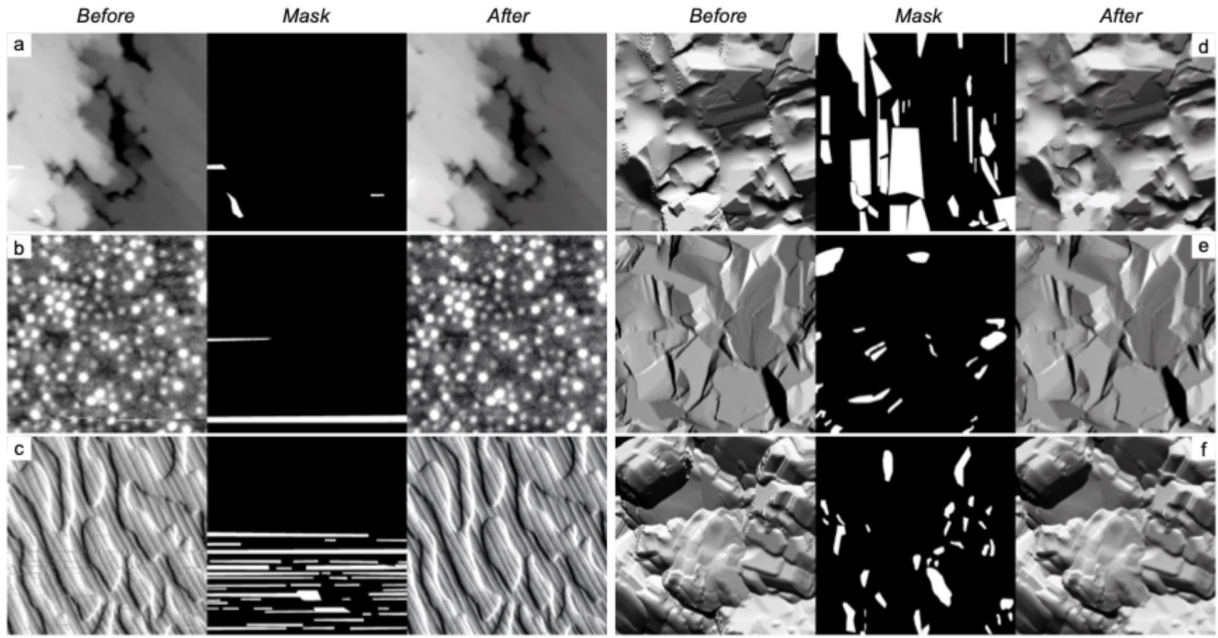

**Figure 3.** Qualitative restoration examples on authentic SPM datasets. Examples of LoRA-adapted BrushNet performance on realistic SPM artifacts, including (a–c) structured horizontal line-scan dropouts and (d–f) irregular, large-area gain-related artifacts. Each sub-panel set shows the original corrupted micrograph ("Before"), manually refined masks derived from SAM-generated segmentation ("Mask"), and corresponding restored images produced by the model ("After"). Images demonstrate restoration across diverse SPM samples: (a) An metal–organic framework ($CH_3–2NH_2MnH_2PO_2–3$), (b) PVDF–Ag composite, (c) PZN–PT ferroelectric, and (d–f) LAGP solid electrolyte.

**D2. Tip-induced directional tailing artifacts.** The second major artifact mechanism is a directional "tailing" of features caused by the probe tip itself. These artifacts arise from the finite size and interaction of the scanning tip with the sample. On hard, sharp-edged structures, the effect is often due to tip convolution: because an AFM tip has a non-zero radius, sharp features can appear broadened or elongated along the scan direction, effectively lowering the lateral resolution. On softer samples, the tip may actually deform or drag the material as it scans, leading to smeared or stretched features behind the tip. Such tip-induced aberrations are usually more subtle than overt line noise and can easily be misinterpreted as real characteristics of the samples. For instance, a flake that appears with a long tail or doubled outline may be mistaken for a distinct second phase or a domain extending from the main feature. This risk is especially high in the channels such as the amplitude or phase signal, where tip-artifacts can alter the contrast in a way that mimics legitimate material property variations. In Figures 4(a) – (d), we present examples of directional tailing on two disparate samples: a two-dimensional (2-D) $MoS_2$ crystal and a piece of tissue sample from a mouse bone. In the uncorrected images, prominent features exhibit streak-like extensions in the fast scan direction. In the $MoS_2$ sample, the edges of the crystal appear falsely blurred or pulled outward, which could be erroneously interpreted as a gradual phase transition region. Similarly, the bone surface shows elongated bright traces adjoining some structures, potentially misleading one to identify micro-fractures or fibrous outgrowths that are not truly there. After restoration by the model, these spurious tails and halos are removed, recovering the crisp shape of the features. The $MoS_2$ crystal's boundaries, for example, revert to their true sharpness without the faux extension, and the bone's microstructural details appear without the trailing artifacts, clarifying that those elongated patterns were not real features in the samples.

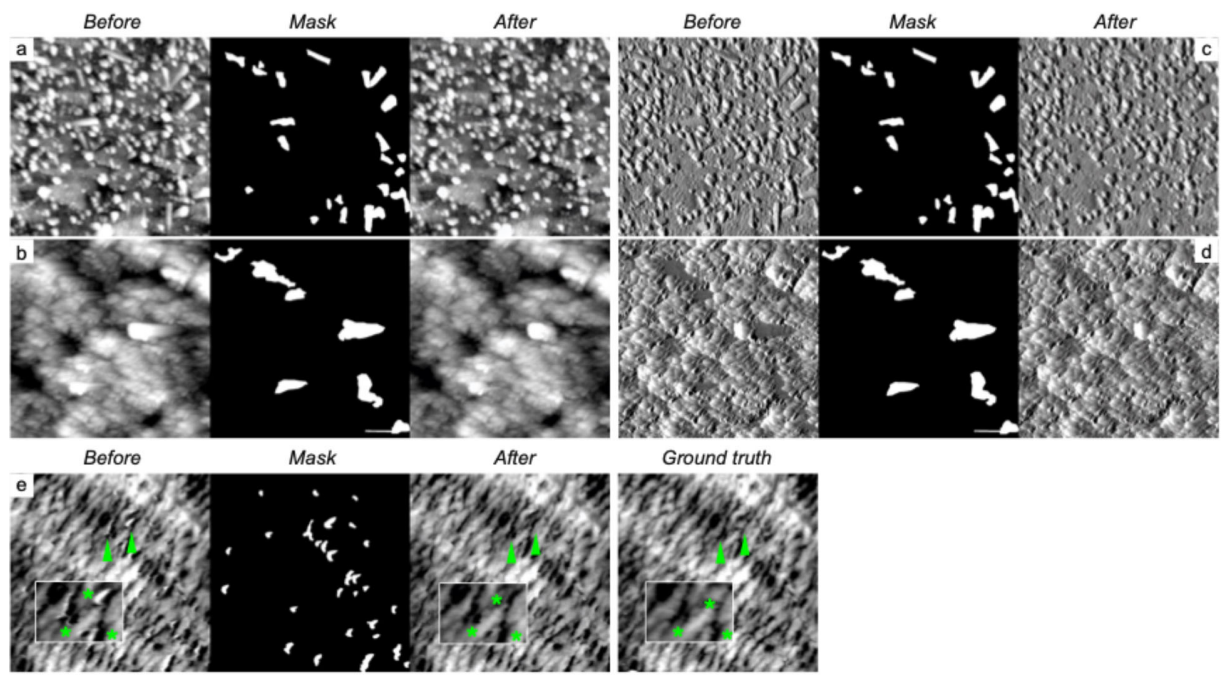

**Figure 4.** Restoration of tip-induced directional tailing artifacts in various SPM samples. Panels (a–d) illustrate directional tailing artifacts on representative SPM samples: (a, b) a 2-D $MoS_2$ crystal and (c, d) mouse bone tissue. Each set displays the original micrograph ("Before"), manually refined masks identifying artifact locations ("Mask"), and corresponding restored images ("After"). Panel (e) highlights tip-induced deformation on a soft PVDF–MXene nanocomposite, where black regions represent natural pores in the PVDF matrix. The original image ("Before") was acquired under conditions that induced deformation and elongation of these pores due to tip dragging. The "Ground truth" image was obtained from multiple low-force scans allowing the sample surface to recover. The restored image ("After") effectively removes the deformation, restoring the accurate, round pore shapes, demonstrating the model's proficiency in correcting subtle, physically-induced artifacts without compromising authentic structural details.

Figure 4(e) further demonstrates artifact correction in a soft PVDF–MXene nanocomposite, where the polymer matrix was physically dragged by the scanning tip, creating visible deformation in the scan direction. Here, black regions correspond to natural pores in the PVDF matrix, rather than embedded particles. The original ("before") image, taken under conditions of relatively high tip force, shows these pores' boundary distorted into the elongated shapes due to tip-induced sample deformation. For validation, repeated low-force scans allowed the polymer to relax and provided a stable ground-truth reference topography. Impressively, the model's single-step restoration closely matches this carefully obtained reference image. After correction, the pores revert to their natural, round shapes without the artificial elongation, effectively reversing the tip-induced surface deformation.

**D3. Phase-hop artifacts in tapping mode.** The final category of artifacts arises from abrupt phase hops and feedback loss events in the tapping-mode AFM. In tapping mode operation, the cantilever is oscillated at or near its resonant frequency, and a stable phase offset (typically on the order of 90°) is maintained between the drive and the cantilever response. When the tip encounters a sudden change in interaction forces, for example, at a sharp sample edge with strong electrostatic or capillary attraction, the cantilever's oscillation can momentarily become unstable. The feedback controller may fail to maintain the tip contact, resulting in a brief excursion of the amplitude and phase. These events create regions of invalid data in the height image (often manifesting as anomalous patches or streaks near the feature that triggered the hop). Importantly, phase-hop artifacts are not random noise superimposed on the signal, but rather the segments of the image where the measured signal is qualitatively wrong, in other words, the AFM effectively is not measuring the surface during those instants. As a result, such artifacts can be hard to recognize from the topography channel alone, at first glance, they may blend into the image and may be misconstrued as real topographic features if one is unaware of the instability. A reliable way to detect these instabilities is by examining the phase channel recorded alongside the topography. At true free oscillation (no tip contact) the cantilever's phase lag approaches ∼90° relative to the drive; hence, pixels where the phase is near ±90° serve as a clear indicator of a tip-sample disengagement or "hop" event. By thresholding the phase image (for instance, marking all pixels beyond an 85–90° phase lag), one can obtain a binary mask highlighting the regions of phase discontinuity due to unstable tip engagement. Figures 5(a) – (f) illustrate the impact of phase-hop artifact removal on two samples (a ferroelectric PVDF film in Figs. 5(a) – (c), and a mixed FeO/CdS composite in Figs.5(d) – (f)). In the original "raw" images, certain boundary areas contain subtle anomalous contrast that is only evident as artifact when cross-checked with the phase image. At full image scale, these artifact-afflicted areas are not immediately obvious, and one may interpret the spurious contrast as part of the sample's morphology without the guidance of the phase signal. After processing, the model successfully suppresses the phase-hop artifacts in both examples. The model-processed images show uniform, continuous contrast across the previously corrupted regions, with no sign of the discontinuities that are present before. Crucially, the underlying real features, such as domain structures in the PVDF sample, or particles in the FeO/CdS composite, are left intact and clarified. The model effectively inpaints the invalid zones caused by feedback loss, yielding a final image where the true morphology is cleanly revealed and no false features persist.

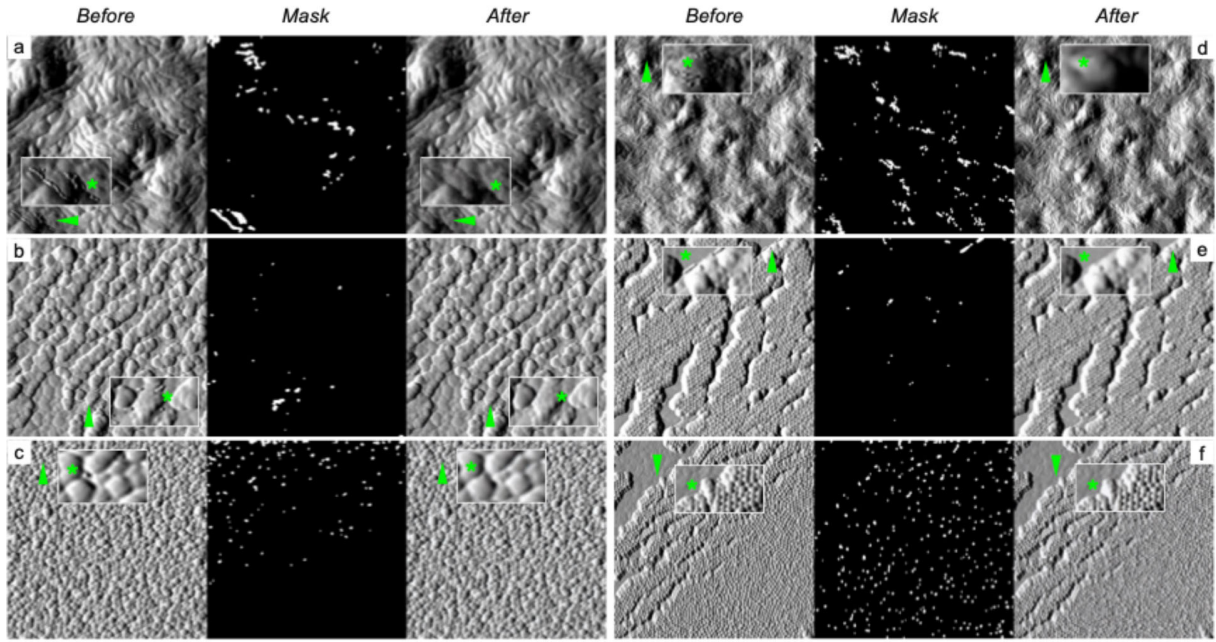

**Figure 5.** Restoration of phase-hop artifacts in tapping-mode AFM images. Panels (a–f) demonstrate effective restoration of phase-hop artifacts for two representative samples measured in tapping-mode AFM: (a–c) a ferroelectric PVDF film and (d–f) a mixed FeO/CdS composite. Each set presents the original micrographs containing subtle anomalous contrast areas due to phase-hop events ("Original"), binary masks identifying artifact regions derived from thresholding the corresponding phase channel ("Mask"), and the restored images generated by the LoRA-adapted BrushNet model ("model-processed").

DISCUSSION

**A. Prompt effectiveness and necessity in SPM image restoration.**

In diffusion-based generative models, textual prompts typically serve to guide image synthesis by providing semantic conditioning through a pretrained text encoder (such as CLIP[30], Contrastive Language–Image Pre-Training). In principle, detailed textual descriptions can specify material types, scan resolutions, and imaging channels, potentially improving restoration by aligning model priors with domain-specific knowledge. However, whether such textual conditioning translates into tangible performance gains in specialized domains, particularly for SPM images, has remained untested.

To explicitly address this gap, we conducted a systematic ablation study to quantify the effect of textual prompts on restoration performance using our LoRA-adapted BrushNet model. Specifically, we compared three distinct textual prompt conditions: (1) a null prompt ("", no semantic guidance); (2) a minimal generic prompt ("*grayscale image*", employed during LoRA fine-tuning); and (3) a full domain-specific prompt ("*a grayscale SPM image of PVDF material at 20μm×20μm, capturing Height channel*"). Quantitative results are summarized in Figure 6 and Table 4 (Statistically significant results ($p < 0.05$) are highlighted in bold.).

**Table 4.** Statistical comparison of prompt effectiveness on PSNR and LPIPS metrics (n=415 paired samples).

| Prompt Comparison | ΔPSNR (dB) | *p(PSNR)* | Cohen's *d* | ΔLPIPS | *p*(LPIPS) | Cohen's *d* |
|---|---|---|---|---|---|---|
| "Null" vs. "Generic" | **+0.29** | **1.13e-7** | 0.265 | +0.000001 | 0.994 | -0.0004 |
| "Null" vs. "Full" | –0.17 | **0.018** | –0.117 | +0.000841 | **3.24e-13** | 0.3696 |
| "Generic" vs. "Full" | –0.45 | **9.88e-10** | –0.3071 | +0.000842 | **1.24e-13** | 0.3765 |

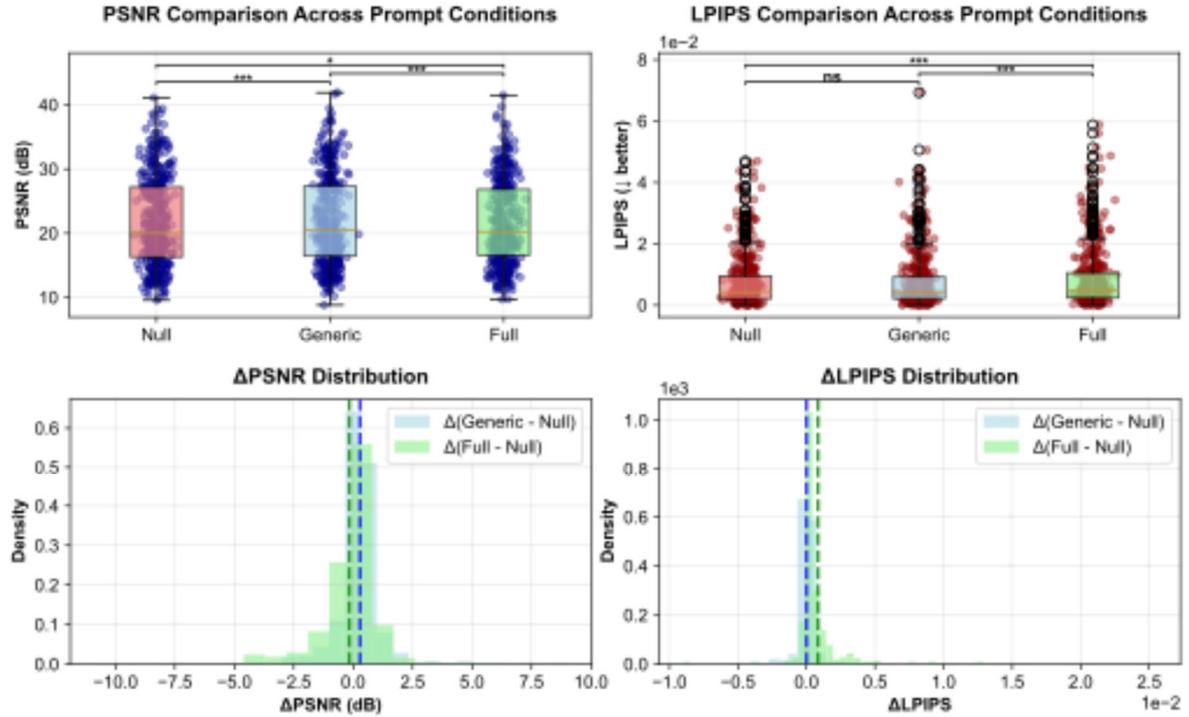

**Figure 6.** Prompt-level ablation study on pixel fidelity and perceptual similarity. Comparison of PSNR (left) and LPIPS (right) across three prompt conditions: null prompt, generic prompt, and a full domain-specific prompt. Boxplots show per-image metrics over 415 benchmark masks, with paired t-test significance levels denoted as: ***$p < 0.001$, **$p < 0.01$, *$p < 0.05$, and ns otherwise. ΔPSNR and ΔLPIPS histograms (bottom) reveal that domain-specific prompts consistently degrade output quality. PSNR increases marginally under the generic prompt (+0.29 dB), whereas perceptual similarity (LPIPS) remains statistically unchanged. Full prompts result in measurable degradation across both metrics.

It is observed that a minimal generic prompt ("grayscale image") yielded a small yet statistically significant improvement in pixel-level accuracy (ΔPSNR = +0.29 dB, $p < 10^{-7}$) compared to that of the null prompt, while perceptual similarity (LPIPS) remained unaffected ( Δ

LPIPS ≈ 0, p = 0.994). Surprisingly, detailed, domain-specific prompts resulted in statistically significant deterioration in both metrics relative to the generic prompt (∆PSNR = −0.45 dB, p ≈ $10^{-9}$; ∆LPIPS = +0.00084, p ≈ $10^{-13}$). These results clearly indicate that detailed semantic descriptions fail to enhance, and in fact negatively affect, the restoration quality.

**A1. Why do prompts fail in SPM restoration?** The negative impact of detailed prompts is rooted in a semantic mismatch between the CLIP text encoder and the UNet-based generator. CLIP, pre-trained on natural RGB imagery, can only partly parse tokens such as "PVDF–MXene." The UNet, lacking corresponding microscopic priors, cannot translate these text embeddings into meaningful nanoscale structure, so the additional semantic signal acts as noise and reduces fidelity.

**A2. Practical implications.** In principle, this gap could be closed by jointly retraining both the text encoder and the generator on a large domain-specific corpus. However, our entire curated dataset (∼5 GB, 7,390 frames) is orders of magnitude smaller than the terabyte-scale corpora required for stable co-training. Moreover, describing masked regions with sufficient precision ("grain boundary," "slip band") would demand labor-intensive expert annotation and might still introduce hallucinated detail.

**A3. Recommended strategy.** We therefore retain a single generic prompt ("grayscale image") during LoRA fine-tuning, mainly to stabilize tonal range and disable textual conditioning entirely at inference. This design simplifies deployment, halves GPU memory during evaluation, and, as showed by Table 4 and Figure 6, avoids the semantic confusion that degrades restoration quality.

**B. Classical baselines versus a LoRA-adapted BrushNet.**

To compare our LoRA-adapted BrushNet with the established practices, we compared it with five non-learning baselines that remain common in the microscopy workflows: biharmonic spline interpolation, Navier–Stokes (NS) diffusion, PatchMatch exemplar copying, polynomial surface fitting, and Telea fast-marching inpainting. A single binary mask was applied to 20 representative frames randomly drawn from the 415-image benchmark, providing a rapid yet diverse validation set. Four complementary metrics: PSNR, LPIPS, MSE and SSIM, capture both pixel accuracy and perceptual fidelity.

**B1. Quantitative performance.** Table 5 shows that LoRA outperforms every classical alternative. Relative to the strongest baseline, LPIPS drops upto four-fold (31.0 → 7.6 × 10⁻³) and PSNR rises by ~4.7 dB, while SSIM reaches 97.8 %. These gains indicate that the data-driven priors learned during LoRA fine-tuning capture the multi-scale texture of SPM images that analytic interpolators cannot reproduce.

**Table 5.** Quantitative comparison of inpainting methods on a 20-image validation subset.

| Method | $LPIPS_{\times 10^3}$ ↓ | PSNR↑ | $MSE_{\times 10^3}$ ↓ | $SSIM_{\times 10^2}$ ↑ |
|---|---|---|---|---|
| **Biharmonic[13]** | **26.48** | **18.88** | **21.67** | **95.93** |
| **NS[14]** | **31.06** | **17.29** | **24.55** | **95.15** |
| Patchmatch[15] | 42.07 | 16.65 | 29.82 | 94.78 |
| Surface fitting[16] | 32.16 | 17.19 | 25.63 | 95.23 |
| Telea[17] | 31.60 | 17.12 | 25.28 | 95.13 |
| **LoRA (Ours)** | **7.64** | **21.96** | **15.80** | **97.80** |

**B2. Qualitative fidelity.** Figure 7 displays four typical scenes. The two best classical solvers: NS diffusion and biharmonic spline interpolation, consistently oversmooth fine detail, erasing fracture edges (row (a)) and nanoparticle granularity (rows (c) and (d)). In contrast, LoRA reinstates sharp crack boundaries, preserves streak orientation without ringing (row (b)), and reconstructs in close agreement with the ground truth.

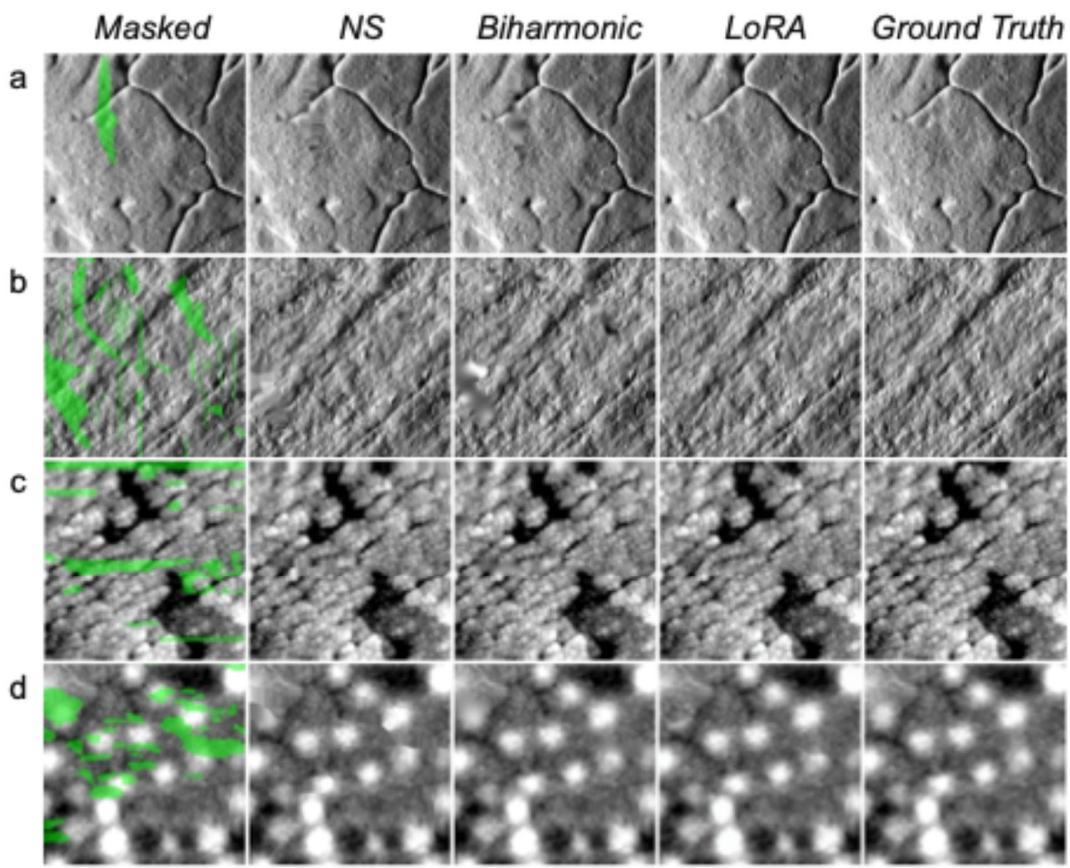

**Figure 7.** Visual comparison of restoration quality on four SPM motifs. Columns (left to right): input with mask (green overlay), Navier–Stokes diffusion, biharmonic spline interpolation, LoRA-adapted BrushNet, and ground-truth reference. Rows (a–d): (a) stainless steel powder (316L); (b) a piece of mouse bone tissue; (c) a mixed FeO/CdS composite; (d) a PVDF–Ag composite.

**B3. Why LoRA, not DreamBooth or ControlNet.** Deep personalized fine-tuning (DreamBooth) and structure-conditioned generation (ControlNet) excel when vast, richly annotated corpora exist or when a precise structural guide is available. However, neither prerequisite holds for Scanning Probe Microscopy images: high-resolution SPM frames remain scarce across the entire community, which are well below the terabyte-scale data typically required for full-model adaptation and genuine artefacts obscure the very edges, poses or segmentation maps that ControlNet would demand as control signals. On the contrary, LoRA can sidestep both limitations. By training <1 % additional low-rank parameters, LoRA adapts BrushNet to SPM texture on a single GPU in under two hours, yields a compact 8 GB adapter, and leaves the backbone's broad visual prior intact. This balance of efficiency and generalization underlies the superior scores and sharper reconstructions reported here.

Collectively, this work demonstrates that a lightweight, low-rank adaptation offers the most practical and accurate route for scientific-image restoration, eclipsing both physics-based interpolators and heavyweight generative fine-tuning strategies.

CONCLUSION

This work shows that minimalist fine-tuning can transform a generic diffusion backbone into a specialist restorer for scientific imagery. Injecting rank-constrained LoRA adapters into BrushNet preserves the rich prior of Stable Diffusion while steering the network toward SPM-specific texture on a single workstation GPU. Quantitative gains on SPM-InpBench translate into qualitative improvements on authentic laboratory data, rescuing frames that would otherwise be discarded.

Beyond immediate benefits for surface-science workflows, two broader insights emerge. First, domain adaptation for diffusion models can succeed without full-network retraining or elaborate prompt engineering; targeted low-rank updates suffice when paired with reliable masks. Second, community-driven resources are pivotal. We therefore envisage building an open SPM semantic library: curated prompts, phase–height pairs and physics-informed priors, to unlock tasks such as joint adapter tuning of the text encoder, principled incorporation of probe–sample physics and extension to exotic SPM modalities. We believe that the code, data and evaluation scripts will benifit this community effort and advance diffusion-based restoration across the microscopy field.

METHODS

This section details the construction of the training corpus, the mask-generation pipeline, the LoRA-adapted BrushNet architecture, the loss and ignore-region strategy, the training schedule, and the evaluation metrics used throughout the study.

**A. Dataset and annotation.**

A total of 1023 grey-scale SPM frames comprising height, amplitude, and phase channels were collected from a commercial SPM system (MFP-3D, Asylum Research, Oxford Instruments, CA, USA), spanning scan sizes from 0.5 μm to 90 μm at resolution of either 512x512 or 1024x1024 pixels. The dataset is partitioned into:

(a) Training subset (*SPM-InpTrain*): 739 frames that serve as inputs for the in-painting network.

(b) Benchmark subset (*SPM-InpBench*): 415 frames held out for quantitative evaluation.

Among all annotated frames, 454 contain naturally occurring artefacts and are further split into: (a) 214 heavy-artefact frames ($\geq$ 2.5 % area), used solely as a source of real noise patches; (b) 240 light-artefact frames (< 2.5 % area), which still provide valid background textures.

From the training subset, we derive 7,390 artefact–clean image pairs via random sampling (mean $\pm$ s.d. = 10.0 $\pm$ 3.1 masks per frame), each indicating a plausible corruption site for inpainting.

**B. Mask-acquisition pipeline.**

We employed a zero-shot segmentation pipeline based on the pretrained APL-SAM model to isolate artefact regions.[31] APL-SAM stands for Adaptive Prompt Learning with Segment Anything Model in which has been described in earlier work. APL-SAM can generate smooth, sub-pixel-accurate masks that align well with the boundary requirements of diffusion-based synthesis. No fine-tuning was performed; instead, masks were predicted directly from raw input frames (Table 6). To ensure high precision, each predicted mask was subjected to rapid expert review and filtered by physical heuristics.

Table 6. Three-stage pipeline that generates a library of artefact masks.

| Step | Action | Key setting | Output |
| --- | --- | --- | --- |
| 1. Zero-shot segmentation | Apply pretrained APL-SAM to noisy frames | Input size $256^2$; no fine-tuning | Coarse masks with smooth boundaries |
| 2. Expert refinement | Interactive correction via web interface (~5 s per mask) | — | 7467 manually verified masks (mean $\pm$ s.d. = 10.0 $\pm$ 3.1 masks per frame) |
| 3. Physics filter | Discard masks with negligible surface contrast | $\Delta h$ < 0.2 nm | Final accepted masks used for training |

**C. Model and Training.**

A frozen Stable-Diffusion v1.5 UNet provides the generative prior;[32] its backbone is a stack of ResNet blocks interleaved with cross-attention layers that inject text conditioning.[33] BrushNet duplicates this UNet but strips out every cross-attention (Spatial-Transformer) module, leaving only the original ResNet down-, mid-, and up-blocks to process pure image features. To supply those features, a lightweight BrushNet branch with eleven depth-wise Conv2d adapters (down-blocks → mid-block → up-blocks) ingests a 9-channel tensor comprising: (i) the current noisy latent, (ii) the VAE-encoded masked latent, and (iii) a binary mask. After each stage this branch injects its activations into the corresponding frozen ResNet block through a zero-convolution gate, scaled by a preservation factor w = 0.8. To keep the learnable footprint below 1% of the full model, we adopt LoRA and insert rank-8 adapters at 17 carefully chosen convolutional layers:

(a) Early context capture: down_blocks.0/1 (first two ResNet groups, four convs);

(b) Global mixing: mid_block.resnets.0/1 (four convs);

(c) Late detail refinement: up_blocks.0/1/3 (three shortcut + two convs); and

(d) Cross-branch fusion hot-spots: brushnet_down_blocks.4 and brushnet_up_blocks.10.

Only these LoRA matrices are trained; all original weights remain fixed. This arrangement lets the adapters steer information flow in terms of early low-level edges, mid-level semantic mixing, and late high-frequency detail, while the zero-convolution gates ensure the masked-image guidance enters at every resolution.

**C1. Loss function and ignore-region strategy.** Each "clean" frame is paired with N=10 randomly cropped artefact masks (mean value consistent with Table 6. Two branches are generated on-the-fly:

(a) Fully supervised: artefact patches replace the corresponding region, yielding a perfect clean target.

(b) Light-artefact: a binary ignore mask is produced by thresholding the natural artefact mask. Pixels inside the ignore mask are excluded from the loss.

In this work, we denoted the real-artefact pixels as $A \subset 1, \ldots, H \times 1, \ldots, W$, the synthetic in-painting mask by $M$, and the evaluation set by $\Omega = (\neg M) \cup (M \setminus A)$. With predicted noise $\hat{\epsilon}_\theta$ and ground truth $\epsilon$, the training loss is therefore defined as:

$$\mathcal{L}(\theta) = \frac{1}{|\Omega|} \sum_{p \in \Omega} \|\hat{\epsilon}_\theta(p) - \epsilon(p)\|_2^2.$$

This dual-track design conserves contextual integrity, suppresses bias from residual artefacts, and maximizes the utility of a small corpus.

**C2. Training schedule.** Two training regimes are employed (Table 7): Aligned-batch (algorithmic fairness) and Throughput-saturated (hardware-limited maximum). Aligned-batch keeps micro-batch = 1 and effective batch = 4 to match gradient-noise scale; Throughput-saturated maximizes per-setup utilization under the 24 GB/card budget. In both, all Stable-Diffusion UNet weights are frozen and LoRA adapters (rank = 8) are inserted only into the BrushNet branch. Checkpoints are saved every 0.2–1k steps; the best validation checkpoint is selected post hoc by peak PSNR (ties broken by lowest LPIPS) computed inside the green masks.

Table 7. Full BrushNet retraining vs. LoRA fine-tuning.

| Setting | Full BrushNet retraining | LoRA fine-tuning (rank = 8) |
|---|---|---|
| GPUs | 4 × NVIDIA A5000 (24 GB) | 1 × NVIDIA A5000 (24 GB) |
| Trainable parameters | 619.2 million | 0.7475 million (0.1206%) |
| Initial learning rate (LR) | $1.5 \times 10^{-5}$ | $2 \times 10^{-6}$ |
| Optimizer | AdamW, $\beta_1 = 0.9$, $\beta_2 = 0.999$ | AdamW, $\beta_1 = 0.9$, $\beta_2 = 0.999$ |
| LR schedule | Constant + warm-up | Cosine decay + warm-up |
| **Aligned-batch** | | |
| micro-batch | 1 | 1 |
| accumulation | 1 | 4 |
| effective batch[a] | 4 | 4 |
| **Throughput-saturated** | | |
| micro-batch | 3 | 12 |
| accumulation | 1 | 1 |
| effective batch | 12 | 12 |

a) Effective batch (= #GPU × micro-batch × accumulation)

## D. Evaluation metrics.

SPM images are inherently monochrome, contain instrument-specific textures, and lack semantic or color priors. Evaluation metrics developed for natural RGB imagery, such as ImageReward,[34] Aesthetic Score,[35] Human Preference Score v2 (HPS),[36] and CLIP-based similarity, were found to be unreliable for SPM image analysis. These metrics rely heavily on color semantics or prompt-image alignment, which are absent in the SPM images. We therefore adopt two complementary categories of metrics to evaluate restoration performance.

**D1. Pixel-level Accuracy.** We report three metrics that assess reconstruction fidelity at the pixel level:

(a) PSNR quantifies the ratio between the maximum possible signal and the noise introduced by reconstruction;

(b) MSE directly measures pixel-wise deviation between the restored region and the ground truth; and

(c) SSIM evaluates local structural consistency by comparing luminance, contrast, and structure between patches.

**D2. Human-aligned Quality.** To complement pixel-wise metrics, we use LPIPS, which computes distance in a deep feature space extracted from a pretrained convolutional network (e.g., VGG). LPIPS is known to correlate well with human perception of texture and structure, especially in grayscale or domain-specific contexts.

In this study, all metrics are computed solely within the inpainted (masked) region using the known ground truth as reference, and scores are normalized by the number of masked pixels.

ASSOCIATED CONTENT

**Supporting Information**.

The following files are available upon requests.

Supporting Information – Additional figures, inpainting results on diverse samples, ablation study tables, and description of ignore-region masks.

Supplementary Videos – mp4 videos linking image IDs with sample origin, acquisition settings, and mask segmentation indices.

Dataset availability: The training and benchmark datasets used in this study will be made publicly available upon publication via an open-access repository.

## AUTHOR INFORMATION

**Corresponding Author**

Prof. Kaiyang Zeng – Department of Mechanical Engineering, National University of Singapore, 9 Engineering Drive 1, 117576, Singapore; NUS Research Institute (NUSRI), No. 377 Linquan Street, Suzhou Industrial Park, Suzhou, Jiangsu Province, 215123, China; E-mail: mpezk@nus.edu.sg.

**Author Contributions**

Z. Wei conceived the study, developed the methodology, and carried out all experiments and analyses. K. Zeng supervised the project. Y.Shen developed and fine-tuned APL-SAM for this work. W. Lu contributed the initial dataset and domain expertise in materials imaging. G. Ho supervised the data collection process. All authors reviewed and approved the final manuscript.

**Funding Sources**

This work was supported by the Ministry of Education (MOE), Singapore, through the National University of Singapore (NUS) under the Academic Research Funds (No. A-0009122-01-00). Z. Wei acknowledges support from the NUS Research Scholarship (Industry Relevant) for NUS Ph.D programme..

**Notes**

The authors declare no competing financial interest.

ACKNOWLEDGMENT

We would like to thank Mr. Hanlin Chen (Department of Computer Science, NUS), Dr. Qibin Zeng (Institute of Materials Research and Engineering, A*STAR, Singapore), Mr. Guochen Peng, Ms. Shihua He, and Ms. Chunlei Xu (Department of Mechanical Engineering, NUS) for their assistance in providing training data and domain guidance.

ABBREVIATIONS

SPM, scanning probe microscopy;

LoRA, low-rank adaptation;

LPIPS, learned perceptual image patch similarity;

SAM, segment anything model.